# Discovering Multiple Constraints that are Frequently Approximately Satisfied

**Geoffrey E. Hinton**
Gatsby Computational Neuroscience Unit
University College London
17 Queen Square
London WC1N 3AR
ENGLAND

**Yee-Whye Teh**
Department of Computer Science
University of Toronto
10 Kings College Road
Toronto Ontario M5S 3G4
CANADA

## Abstract

Some high-dimensional datasets can be modelled by assuming that there are many different linear constraints, each of which is Frequently Approximately Satisfied (FAS) by the data. The probability of a data vector under the model is then proportional to the product of the probabilities of its constraint violations. We describe three methods of learning products of constraints using a heavy-tailed probability distribution for the violations.

## 1 CHOOSING A GENERATIVE MODEL

High-dimensional data can be modelled by assuming that it was produced by some type of stochastic generative procedure whose structure and parameters can be estimated from observed data. The first step in this generative approach is to choose a class of generative models and there are several important criteria governing this choice.

- It should be easy to **generate** samples from a fully specified model.

- It should be easy to **evaluate** the log probability of a test case under the generative model.

- It should be easy to **infer** the posterior distribution over unobserved variables in the model when some variables are observed.

- It should be easy to **learn** the parameters (or distributions over the parameters).

- The model class should be flexible enough to **represent** the structure of the data efficiently.

Unfortunately, for data like images or speech it is hard to satisfy all of these criteria simultaneously. Mixture models, for example, satisfy the first four criteria but fail on representational adequacy for data in which several separate causes conspire to produce each observed data vector. If, for example, an image can contain several objects, an accurate mixture model of whole images must have a separate mixture component for every possible *combination* of objects and it is therefore exponentially inefficient.

Directed, acyclic graphical models (*i.e.* causal models) are an efficient way to represent multiple simultaneous causes so they can overcome the representational inadequacies of mixture models, but exact inference becomes intractable when nodes have more than a few parents which is typically the case for sensory data like speech or images.

### 1.1 VARIATIONAL INFERENCE IN CAUSAL MODELS

Progress has recently been made by using approximate inference techniques in directed acyclic graphs that are too densely connected to allow exact inference. The true posterior distribution over the unobserved variables, $P$, is approximated by using a distribution, $Q$, that lies within a simpler and more tractable class of distributions and that minimizes the Kullback-Leibler divergence $KL(Q||P)$. When $Q$ is used in place of $P$ for evaluating the log probability of a test case, it gives a lower bound whose tightness is $KL(Q||P)$. When $Q$ is used in place of $P$ for adjusting parameters there is no guarantee that the log likelihood of the parameters increases, but the log likelihood penalized by the inaccuracy of the approximate inference, $KL(Q||P)$, does increase (Neal and Hinton, 1998; Jordan *et. al.*, 1999).

## 2 PRODUCTS OF EXPERTS

Despite the successes of variational inference for fitting causal models, it is worth considering a different approach in which we abandon the very first criterion, ease of generation, and use generative models for which generation is intractable[1] but inference and learning are easy. For tasks like the interpretation of sensory data, inference and learning are what is important and the stochastic generative model is just a convenient way of defining what inference and learning are achieving, so there is no practical requirement for generation to be easy.

A particularly interesting class of intractable generative models are the "Products of Experts" (PoE) in which the probability distribution over a set of observable (visible) variables is defined as the normalized product of the distributions generated by $m$ separate "experts" where each expert typically consists of a tractable latent variable model that is easy to fit to data. For data that lies in a discrete space, a PoE defines a distribution over data vectors, $\mathbf{d}$ in the following way:

$$p(\mathbf{d}|\theta_1...\theta_m) = \frac{\Pi_j f_j(\mathbf{d}|\theta_j)}{\sum_\mathbf{c} \Pi_j f_j(\mathbf{c}|\theta_j)} \quad (1)$$

where $\theta_j$ is all the parameters of individual model $j$, $f_j(\mathbf{d}|\theta_j)$ is the probability of $\mathbf{d}$ under model $j$ (integrating over the latent variables of model $j$) and $\mathbf{c}$ indexes all possible data vectors (i.e. vectors of states of the visible variables). The major advantage of a product of experts is that exact inference is simple. Given an observed data vector, the latent variables of different experts are conditionally independent so there is no "explaining away".

The obvious way to fit a PoE to data is to follow the gradient of the log likelihood:

$$\frac{\partial \log p(\mathbf{d}|\theta_1...\theta_m)}{\partial \theta_j} = \frac{\partial \log f_j(\mathbf{d}|\theta_j)}{\partial \theta_j}$$
$$- \sum_\mathbf{c} p(\mathbf{c}|\theta_1...\theta_m) \frac{\partial \log f_j(\mathbf{c}|\theta_j)}{\partial \theta_j} \quad (2)$$

The first term on the RHS of equation 2 is tractable if the individual experts correspond to tractable models.

---

[1] Some researchers reserve "generative model" for causal models in which it is easy to generate unbiased samples. In this paper it will be used in the wider sense of any stochastic process that can, in principle, be used to produce observations with a well-defined probability distribution.

Unfortunately, the second term on the RHS of equation 2 involves a sum over all *conceivable* observations so it seems as if the only hope is to get a noisy but fairly unbiased estimate of this term by using Markov Chain Monte Carlo to sample from the space of possible observations with a probability approaching $p(\mathbf{c}|\theta_1...\theta_m)$. Fortunately, Hinton (2001) shows that it is helpful to replace the log likelihood of the observed data with a different objective function called "contrastive divergence" which is much easier to optimize than the log likelihood. This allows a product of experts to be fitted to data efficiently. We shall return to contrastive divergence after considering various types of expert and various other ways of fitting them to data.

## 3 TYPES OF EXPERT

An expert, $j$, is simply a way of associating an energy contribution, $E_j(\mathbf{d})$ with any possible data vector, $\mathbf{d}$. The probability of $\mathbf{d}$ under the PoE generative model is then defined to be proportional to $\exp(-\sum_j E_j(\mathbf{d}))$. In a PoE, $E_j(\mathbf{d})$ does not depend on the behaviour of the other experts. More precisely, given the learned parameters of one expert, the energy contribution that it assigns to a data vector is independent of the other experts. The experts typically correspond to simple latent variable models and the energy contributed by expert $j$ then corresponds to $-\log p(\mathbf{d}|\theta_j)$, where $\theta_j$ denotes the parameters of expert $j$.

Within the broad class of energy-based models a lot of attention has been given to models with binary variables and quadratic energy functions (Hopfield, 1981; Hinton and Sejnowski 1986), possibly because such models are familiar to physicists. Such models can be restricted by forbidding connections between hidden units (Smolensky, 1986; Freund and Haussler, 1992) and they then correspond to PoE's in which each hidden unit and its connections constitute an expert (Hinton, 2001) and the energy contribution of an expert corresponds to a free energy (i.e the negative log of the probability with the state of the binary hidden variable integrated out).

The general idea of modelling a probability distribution over visible variables by using an additive energy function does not require binary variables or quadratic energy functions. An interesting alternative is to use real-valued hidden variables whose scalar values are a non-stochastic parameterized function of the real-valued visible variables. The value of a hidden variable is interpreted as the violation of a constraint and the energy contributed by each hidden variable is the negative log probability of this violation under some probability distribution over violations.

An interesting special case of this approach is obtained by using violations that are a linear function of the visible variables. If the cost function for violations, $v$, is of the form $kv^2$ (corresponding to an assumed zero-mean Gaussian distribution) then the strongest linear constraints correspond to the eigenvectors of the covariance matrix with the *smallest* eigenvalues and it is relatively straightforward to find these "minor components". If, however, the assumed distributions of the violations are non-Gaussian, the product of linear constraints model becomes more powerful but harder to fit to data. Consider a single expert that uses a student-t distribution for the violation, $v$,

$$-\log p(v) = \log(1 + kv^2) + c \qquad (3)$$

This cost function is very tolerant of making a very large violation even larger, but very determined to make a fairly small violation even smaller. It therefore corresponds well to the idea of a Frequently Approximately Satisfied constraint that is occasionally strongly violated.

## 4   A SIMPLISTIC WAY OF FITTING CONSTRAINTS

For the cost function in equation 3 with the value of $k$ fixed, it is easy to fit *one* linear FAS constraint of the form

$$v(\mathbf{d}) = \sum_i w_i d_i \qquad (4)$$

where $w_i$ is a learned weight and $d_i$ is the $i^{th}$ component of data vector $\mathbf{d}$. We simply alternate between steps that rescale the weights to satisfy the constraint $\sum_i w_i = 1$ and gradient descent steps in the the cost function

$$E_{total} = \sum_{cases\, c} \log(1 + kv(\mathbf{d}^c)^2) \qquad (5)$$

Even this very simple learning rule discovers interesting contraints. Figure 1 consists of synthetic images of straight edges with random positions, orientations and intensities. High-frequency anticorrelated noise has been added to make it harder to predict the intensity of a pixel exactly from its neighbors. When the simple learning rule is applied using a circular window on these images, it produces the gratings shown in figure 2. A grating is the best filter for **ignoring** an edge of unknown location and orientation because the grating will produce an output close to zero unless the edge is in almost the same orientation as the grating in which case the output may be far from zero. The idea that an oriented filter is ideal for ignoring edges comes as a surprise to some vision researchers. If we implemnt a version of minor components analysis by using a parabolic cost function for the violations, the filters become circularly symmetric.

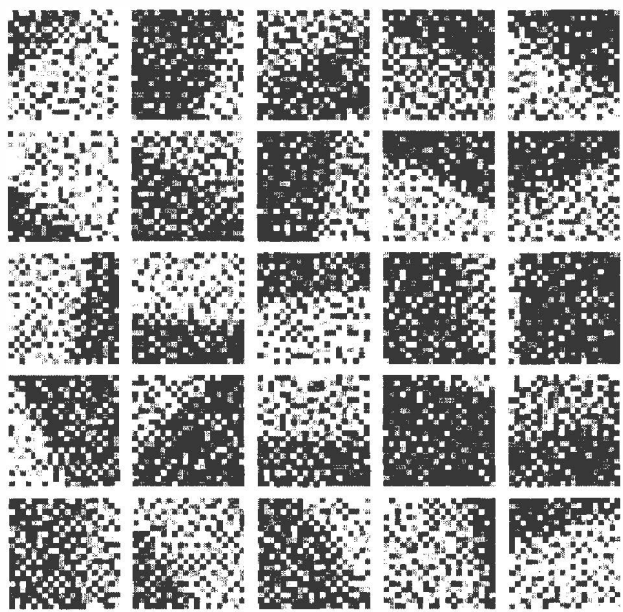

Figure 1: Synthetic edge images are generated by choosing a random orientation and position, a random low intensity in the range $[0, 0.3]$, and a random high intensity in the range $[0.7, 1]$. The hard edge is then softened by blending the high and low intensities using the logistic function $\sigma(d) = 1/(1 + \exp(-2d))$ where d is the distance in pixels of the center of a pixel from the edge. Finally, high-frequency anti-correlated noise is added. At each pixel, we sample $x$ from a zero-mean, unit-variance Gaussian and then add $0.4x$ to the pixel and subtract $0.1x$ from each of its four neighbors.

To encourage the gratings in figure 2 to be different from one another we used a simple heuristic that was weakly inspired by research on boosting (Freund and Schapire, 1995). In boosting, experts are learned sequentially and the data used to train one expert is obtained by reweighting the training data so that data that is not well modelled by the previous experts is given a high weight. Our experts are density models as opposed to conditional density models, we train them all in parallel, not sequentially, and we reweight each training case by its total violation energy, $E$, rather than $\exp(E)$, but apart from that it is like boosting. Reweighting the data tends to make the experts differ-ent because it causes an uncommitted expert to learn faster on high energy cases that are far from the constraint planes that have been learned by the other experts. The uncommitted expert is therefore more strongly attracted to FAS constraints that have not already been found by other experts.

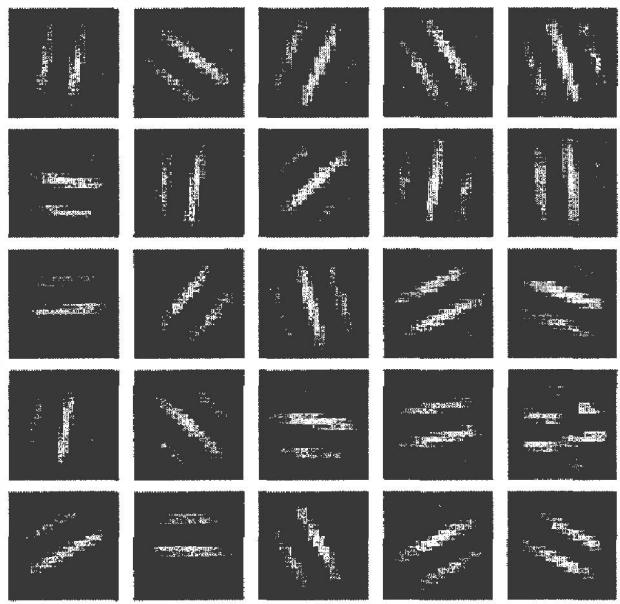

Figure 2: 25 constraints that were learned on images of the type shown in figure 1. Each constraint is a linear filter whose output is typically close to zero. White represents positive filter weights, black negative, and gray is zero. The filters are all global because the images have strong long-range correlations. The constant $k$ in Eq. 5 was fixed at 100. The weights are updated 4000 times using $10^{-7}$ times the gradient of $E_{total}$ on a batch of 1000 images plus 0.98 times the previous weight update. When many filters are learned at the same time, the gradient on each case is also weighted by the total violation energy on that case divided by the average total violation energy over all cases in the batch.

It seems quite likely that applying boosting properly would give better results. The fact that the normalization term in Eq. 1 cannot be computed does not matter because reweighting the training data only requires the relative probabilities of the training cases. Reweighting the training data by the reciprocal of its probability under the previous experts can be very problematic for high-dimensional density models because the probability of one particular training case is often much smaller than all the others, so the worst case dominates the reweighted training set. But this objection may be much less serious for experts that use a heavy-tailed distribution over a scalar variable.

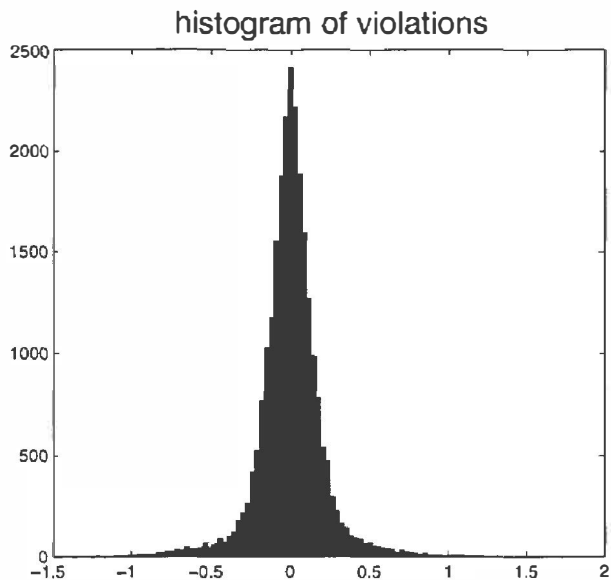

Figure 3: A histogram of outputs of the filters in figure 2 on a batch of 1000 images.

However, there does not seem to be anything to prevent boosting from learning a set of FAS constraints that give much higher probability to some unobserved region of the dataspace than to the training data. To avoid this it is necessary to take into account the second term in equation 2 or some surrogate for this term.

## 5   FITTING MANY CONSTRAINTS USING PSEUDO-LIKELIHOOD

Given the parameters of the experts in a PoE, it is easy to compute the energy of any actual or possible state of all the visible variables. It is therefore easy to do Gibbs sampling by considering the visible variables one at a time and picking a new value for the chosen visible variable, $i$, from its posterior distribution given the current values of the other visible variables. This posterior distribution is proportional to $\exp(-E(\mathbf{d}_{\bar{i}}, d_i=a))$ for each possible state, $a$, of the visible variable $d_i$. By repeatedly updating visible variables until a stationary distribution is reached, it is possible to sample visible states from the generative distribution defined by the PoE. Unfortunately this can be very computationally expensive and it only gives a noisy estimate of the gradient of the log likelihood defined by equation 2.

Instead of attempting to maximize the log likelihood of the observed data, we could attempt to maximize the log pseudo-likelihood (Besag, 1975) which is defined

as:

$$\sum_i \sum_c \log p(d_i^c | \mathbf{d}_{\bar{i}}^c) \quad (6)$$

where $d_i^c$ is the state of visible variable $i$ on training case $c$ and $\mathbf{d}_{\bar{i}}^c$ denotes the states of all the other visible variables. So long as we maximize the pseudo-likelihood by learning the parameters of a single global energy function, the conditional density models for each visible variable given the others are guaranteed to be consistent with one another, so we avoid the problems that can arise when we learn $n$ separate conditional density models for predicting the $n$ visible variables.

Rather than using Gibbs sampling to sample from the stationary distribution, we are learning to get the individual moves of a Gibbs sampler correct by assuming that the observed data is from the stationary distribution so that the state of a visible variable is an unbiased sample from its posterior distribution given the states of the other visible variables. If we can find an energy function that gets the individual moves correct, there is no need to ever compute the gradient of the log likelihood. Pseudo-likelihood therefore replaces the exponentially expensive partition function in Eq. 1 with a one-dimensional partition function for each visible variable, and this makes it feasible to compute exact derivatives which allows optimization methods like conjugate gradient to be used.

To get the derivative of the log pseudo-likelihood on a given training case, we first compute the energy contributed by each of the $m$ constraints given the observed states of all $n$ visible variables. This requires $O(mn)$ operations. Then, assuming the visible variables are discrete or quantized, for each of the $a$ alternative states of each visible variable we compute how the energy contributed by each constraint would vary if the state of that visible variable was changed. So computing the exact derivative of Eq. 6 requires only $O(mna)$ operations per training case.

Empirical research on the effectiveness of finding FAS constraints by optimizing the log pseudo-likelihood is underway but has not yet produced any impressive results and it appears to be considerably slower than the method which we describe next.

## 6 FITTING MANY CONSTRAINTS USING CONTRASTIVE DIVERGENCE

Consider an expert, $j$, that has a binary latent variable, $s_j$, that chooses between two different, zero-mean Gaussian models for the violation of a linear constraint. The parameters of the expert are the weights $\lambda_j$ that define the constraint plane, the variances $\sigma_{j1}^2, \sigma_{j0}^2$ and $m_j$, where the probability of choosing variance $\sigma_{j1}^2$ is $1/(1 + \exp(-m_j))$. By learning to make one Gaussian broad and the other narrow, the expert can implement a heavy-tailed model of the violations that is appropriate for a FAS constraint.

If the latent state of an expert is known, the expert represents an improper distribution that is Gaussian in the direction orthogonal to the constraint plane and uniform in the other $n-1$ orthogonal directions. Assuming the experts represent at least $n$ linearly independent constraints, their product represents a Gaussian distribution in the space of data vectors, so it is possible to sample randomly from this space provided the states of all the binary latent variables, $\mathbf{s}$, are given. It is also straightforward to sample from the states of the binary latent variables given a data vector because they are conditionally independent. So by alternately sampling from $p(\mathbf{s}|\mathbf{d})$ and $p(\mathbf{d}|\mathbf{s})$ it is possible to perform Gibbs sampling and produce samples from $p(\mathbf{c}|\theta_1...\theta_m)$ in Eq. 2. Following the gradient in Eq. 2 maximizes the log likelihood of the data which is equivalent to minimizing the Kullback-Leibler divergence, $KL(P^0||P_\theta^\infty)$ between the observed data distribution $P^0$ and the equilibrium distribution over the visible variables, $P_\theta^\infty$, that is produced by prolonged Gibbs sampling[2].

Hinton (2001) describes an effective learning procedure that is much faster than running the Markov chain to equilibrium. We simply run the chain for one full Gibbs step by sampling $\mathbf{s} \sim p(\mathbf{s}|\mathbf{d})$ and $\widehat{\mathbf{d}} \sim p(\widehat{\mathbf{d}}|\mathbf{s})$, where $\widehat{\mathbf{d}}$ is a one-step reconstruction of $\mathbf{d}$. If we then use $\widehat{\mathbf{d}}$ in place of an equilibrium sample from the generative model, the gradient learning rule defined by Eq. 2 approximates the gradient of the *contrastive divergence*, $KL(P^0||P_\theta^\infty) - KL(P_\theta^1||P_\theta^\infty)$, where $P_\theta^1$ is the distribution of the one-step reconstructions of the observed data.

Because the distribution $P_\theta^1$ is closer to equilibrium than the distribution $P^0$, the contrastive divergence cannot be negative. The main justification for minimizing contrastive divergence is that it is easy to do and it produces good results. But an intuitive justification may also be helpful. If we had a perfect model of the data and we started the Markov chain used in Gibbs sampling at the data distribution it would just sit there going nowhere. Inadequacies in the model show up as a consistent tendency for the Markov chain to move away from the data distribu-tion. We do not need to run the chain all the way to equilibrium to sense this tendency. One step is sufficient. If $KL(P^0\|P_\theta^\infty) > KL(P_\theta^1\|P_\theta^\infty)$ we already know that the model is inadequate and we also have information about how to fix it.

---

[2] $P^0$ is a natural way to denote the data distribution if we imagine starting a Markov chain at the data distribution at time 0.

## 6.1 THE GRADIENTS

For a data vector $\mathbf{d}$, let the violation be $v_j = \lambda_j^T \mathbf{d}$ and $s_{j1} = E[s_j|\mathbf{d}]$ and $s_{j0} = 1 - s_j|\mathbf{d}$. The gradients represented by the first term on the RHS of Eq. 2 are:

$$\frac{\partial \log f_j(\mathbf{d}|\theta_j)}{\partial m_j} = s_{j1} \quad (7)$$

$$\frac{\partial \log f_j(\mathbf{d}|\theta_j)}{\partial \log \sigma_{j1}^2} = s_{j1}(v_j^2/\sigma_{j1}^2 - 1)/2 \quad (8)$$

$$\frac{\partial \log f_j(\mathbf{d}|\theta_j)}{\partial \log \sigma_{j0}^2} = s_{j0}(v_j^2/\sigma_{j0}^2 - 1)/2 \quad (9)$$

$$\frac{\partial \log f_j(\mathbf{d}|\theta_j)}{\partial \lambda_j} = (s_{j1}/\sigma_{j1}^2 + s_{j0}/\sigma_{j0}^2)v_j \mathbf{d} \quad (10)$$

To get the approximate derivative of the contrastive divergence with respect of each of the parameters $m_j$, $\log \sigma_{j1}^2$, $\log \sigma_{j0}^2$ and $\lambda_j$, we subtract from the derivative above the same derivative computed using $\widehat{\mathbf{d}}$ in place of $\mathbf{d}$, $\widehat{s}_{j1}$ in place of $s_{j1}$, $\widehat{s}_{j0}$ in place of $s_{j0}$ and $\widehat{v}_j$ in place of $v_j$, where $\widehat{s}_{j1}$, $\widehat{s}_{j0}$ and $\widehat{v}_j$ are computed using $\widehat{\mathbf{d}}$ in place of $\mathbf{d}$.

## 6.2 HOW THE RECONSTRUCTIONS ARE PRODUCED

To generate the reconstruction $\widehat{\mathbf{d}}$ of a data vector $\mathbf{d}$, we first sample, for each expert $j$, which Gaussian to use from the posterior $s_j \sim p(s_j|\mathbf{d})$, and then reconstruct by sampling $\widehat{\mathbf{d}}$ from a zero mean Gaussian distribution with inverse covariance $\Sigma = \Lambda D \Lambda^T$ where the $j$th column of $\Lambda$ is $\lambda_j$, and $D$ is a diagonal matrix with $D_{jj} = s_j/\sigma_{j1}^2 + (1-s_j)/\sigma_{j0}^2$.

Suppose that $\Lambda$ is invertible. Then $\Sigma$ is invertible too. The naive way of sampling from $\widehat{\mathbf{d}}$ is to use the Cholesky factorization of $\Sigma$. However this is very inefficient since $\Sigma$ is different for each training case and each iteration, and it requires $O(n^3)$ operations to factorize each $\Sigma$, where $n$ is the dimensionality of $\widehat{\mathbf{d}}$. An alternative method is to first sample $\widehat{\mathbf{u}} = \Lambda^T \widehat{\mathbf{d}}$, which has covariance $\Lambda^T \Sigma^{-1} \Lambda = D^{-1}$, and then compute $\widehat{\mathbf{d}} = \Lambda^{-T} \widehat{\mathbf{u}}$. This is more efficient since $D$ is diagonal, and computing $\Lambda^{-T} \widehat{\mathbf{u}}$ requires only $O(n^2)$ operations (with an $O(n^3)$ overhead per iteration, for all training cases).

When $\Lambda$ is not invertible, there are two problems –

the columns of $\Lambda$ might not be linearly independent, and they might not span the vector space (the two problems can occur independently of each other since the number of constraints $m$ can be greater or less than the number of dimensions $n$). We can solve the first problem by increasing the dimensionality of $\mathbf{d}$. In particular, append $m$ zeros to each input vector, and append $\mathbf{m}_j$ to each $\lambda_j$, where $\mathbf{m}_j$ is the $j$th unit vector of length $m$. Now $\{\lambda_j\}$ is independent, but $\lambda_j \mathbf{d}$ stays unchanged so the gradients (7-10) need not be modified.

Now suppose that $\{\lambda_j\}$ is linearly independent, but does not span the space. Let $U$ be the space spanned by $\{\lambda_j\}$ and $V$ its orthogonal complement. Note that the (non-trivial) nullspace of $\Sigma$ is exactly $V$. Decompose $\widehat{\mathbf{d}} = \widehat{\mathbf{d}}_U + \widehat{\mathbf{d}}_V$ where $\widehat{\mathbf{d}}_U \in U$ and $\widehat{\mathbf{d}}_V \in V$. Then $\Sigma$ is positive-definite on $U$ so we can sample $\widehat{\mathbf{d}}_U$, but $\Sigma$ is exactly 0 on $V$, so $\widehat{\mathbf{d}}_V$ has infinite variance. To deal with this, consider $\widehat{\mathbf{d}}_V$ to be Gaussian distributed with zero mean and variance $\rho$, and take $\rho \to \infty$. Now $\widehat{\mathbf{u}} = \Lambda \widehat{\mathbf{d}} = \Lambda \widehat{\mathbf{d}}_U$ still has covariance $D^{-1}$ so we can sample $\widehat{\mathbf{u}}$ first and set $\widehat{\mathbf{d}}_U = \Lambda^\# \widehat{\mathbf{u}}$ where $\Lambda^\#$ is the pseudo-inverse of $\Lambda$. Now integrate out $\widehat{\mathbf{d}}_V$. This will not affect (7-9). For (10), since $\widehat{\mathbf{d}}_V$ is zero mean, $E[v_j \widehat{\mathbf{d}}] = v_j \widehat{\mathbf{d}}_U$ is independent of $\rho$, and taking $\rho \to \infty$ will not affect it either.

In summary, we reconstruct with the following algorithm :

$$s_j \sim p(s_j|\mathbf{d})$$
$$\widehat{\mathbf{u}} \sim Gaussian(0; D^{-1})$$
$$\widehat{\mathbf{d}} = \Lambda^\# \widehat{\mathbf{u}}$$

Figure 4 shows the constraints that are learned when the algorithm above is applied to $16 \times 16$ patches of images of outdoor scenes.

## 7 MORE GENERAL CONSTRAINTS

This paper has considered planar constraints for which the violation is a linear function of the data. This makes it tractable to map a Gaussian distribution over the violations to a Gaussian distribution over the data space, which was useful when producing the reconstructions required for estimating the gradient of the contrastive divergence. The gradient of the log pseudo-likelihood, however, can easily be computed for smooth non-linear constraints, since it only requires the energy contributed by a constraint and its derivative. So each "constraint" could be a feedforward, multilayer neural network with one output unit whose activity represents the violation.

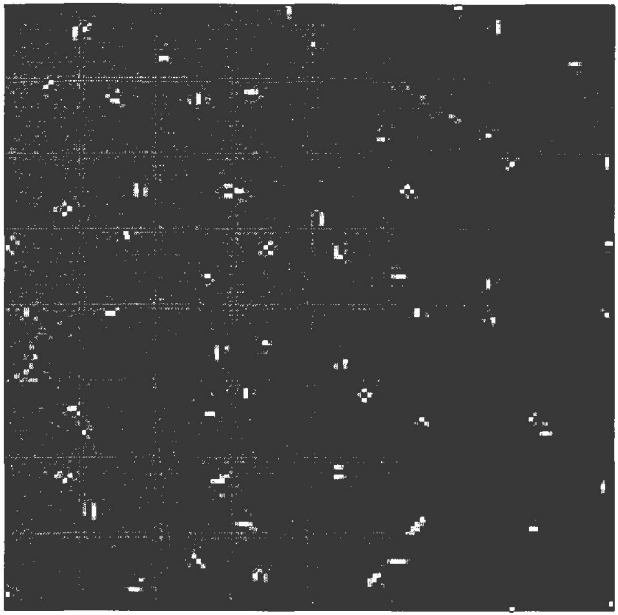

Figure 4: A randomly selected subset of the 256 constraints that were learned on 16 × 16 patches of natural images. Nearly all of the constraints have become quite local and strongly dislike edges or bars in a particular "preferred" orientation. The image patches were preprocessed by adding high-frequency, anti-correlated noise.

## 8 CONSTRAINTS AND NEURONS

Early research on the response properties of individual neurons in visual cortex typically assumed that neurons were rather specific feature detectors that only fired when they found a close match to the feature of interest. For the early stages of the visual cortex, this assumption has largely been replaced by the idea that the receptive fields of neurons represent basis functions and the neural activities represent coefficients on these basis functions. The sensory input is then represented as a weighted linear combination of the basis functions which is equivalent to assuming that the sensory input is generated by a causal linear model with one layer of latent variables and that low-level perception consists of inferring the most likely values of the latent variables given the sensory data. With the added assumption that the latent variables have heavy-tailed distributions, it is possible to learn biologically realistic receptive fields by fitting a linear, causal generative model to patches of natural images (Olshausen and Field, 1996; Bell and Sejnowski, 1996).

However, a linear causal model is just one possible way of interpreting the biological data. It leads to slow, iterative algorithms for computing the activations of the basis functions when they are non-orthogonal, as they must be if they are over-complete. As we have seen, a product of constraints is a very different type of generative model which can give rise to quite similar receptive fields and which allows very rapid inference even when the constraints are highly correlated with one another. This suggests an interesting possibility. Instead of thinking of neurons as devices that are designed to find significant features or to learn basis functions, we can view them as devices that learn to model the constraints in the sensory data and that only complain to higher levels when these constraints are violated. The outputs of neurons are then seen as residual errors, and a multilayer network is seen as a hierarchy of separate models each of which captures the structure remaining in the residual errors from the previous layer. An attractive aspect of this view is that attention generally needs to be directed towards parts of the world that violate our expectations and this is easy to implement if neural activities represent violations of regularities.

### Acknowledgements

We thank Andrew Blake, Peter Green, David Mackay, Guy Mayraz, Sam Roweis, Chris Williams and members of the Gatsby Computational Neuroscience Unit for helpful discussions. This research was funded by the Gatsby Charitable Foundation.